\documentclass[dvipsnames]{article}
\usepackage[T1]{fontenc}
\usepackage{colm2024_conference}
\usepackage{microtype}
\usepackage{graphicx}
\usepackage{wrapfig}
\usepackage{placeins}
\usepackage{needspace}
\usepackage{booktabs}
\usepackage{amsmath}
\usepackage{amssymb}
\usepackage{mathtools}
\usepackage{tabularx}
\usepackage{colortbl}
\usepackage{enumitem}
\usepackage[english]{babel}
\usepackage{tikz}
\usepackage{tcolorbox}
\tcbuselibrary{skins,breakable,listings}
\usetikzlibrary{arrows.meta,positioning}
\definecolor{promptBlue}{HTML}{355070}
\newtcblisting{rubricprompt}[1]{
  enhanced,breakable,listing only,
  colback=white,colframe=promptBlue!80!black,
  colbacktitle=promptBlue!10,coltitle=promptBlue!85!black,
  fonttitle=\bfseries,title={#1},boxrule=0.45pt,arc=1pt,
  left=5pt,right=5pt,top=5pt,bottom=5pt,
  before skip=0.75em,after skip=0.75em,
  listing options={basicstyle=\ttfamily\footnotesize,breaklines=true,
    columns=fullflexible,keepspaces=true,showstringspaces=false}
}
\newcolumntype{L}[1]{>{\raggedright\arraybackslash}p{#1}}
\newcolumntype{Y}{>{\raggedright\arraybackslash}X}
\newcommand{\noindentparagraph}[1]{\par\noindent#1\space}
\newcommand{\Description}[1]{}
\renewcommand{\headrulewidth}{0pt}
\fancypagestyle{firstpage}{%
  \fancyhf{}
  \fancyhead[L]{\includegraphics[height=12mm]{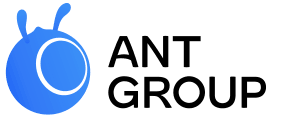}}
  \fancyhead[R]{\includegraphics[height=12mm]{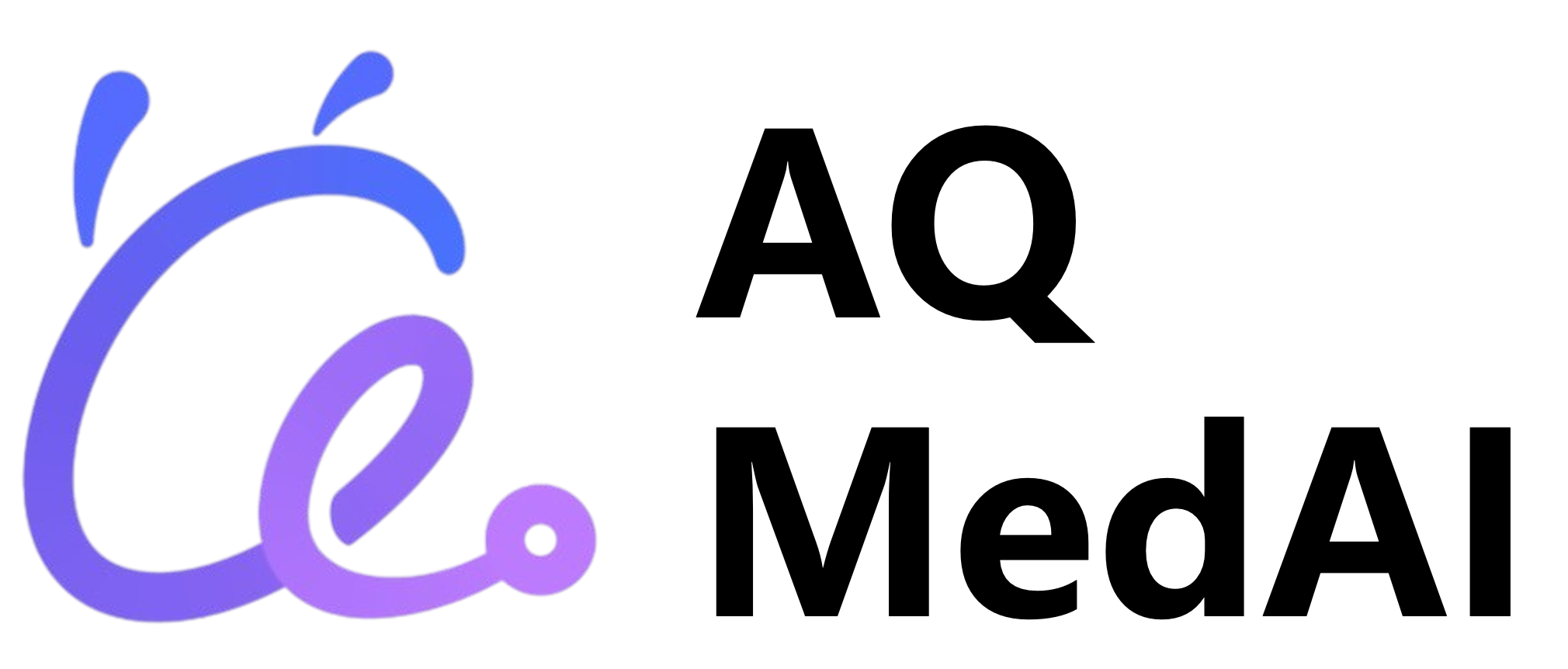}}
  \fancyfoot[C]{\thepage}
  \renewcommand{\headrulewidth}{0.4pt}
}
\title{\textbf{Seed-Evolve-X: Synthesizing Diverse Agentic Tasks through Multi-Strategy Evolution}}
\author{\textbf{Shiyu Liu, Meixiu Long, Dan Yang, Junming Yang, Junjie Wang,}\\
\textbf{Lei Liu, Lin Zhang, Yue Shen, Jian Wang, Jinjie Gu}\\[0.5em]
Ant Group}
\begin{document}
\maketitle

\begin{tcolorbox}[colback=blue!5!white,colframe=blue!20!white,boxrule=2pt,arc=3mm]
\begin{abstract}
Training working agents to perform complex, long-horizon workflows requires diverse, executable tasks and reliable outcome verification. Existing expert-designed synthesis pipelines limit exploration to predefined resources and task patterns. Autonomous strategy generation can broaden exploration, but different strategies may still produce similar scenarios and workflows. As a result, generating more tasks does not necessarily broaden the coverage of the training data.
To address these limitations, we propose Seed-Evolve-X, a framework that creates and evolves task-generation strategies through diversity feedback. Its Multi-Strategy Evolution Harness enables an agent to develop reusable seed-generation skills informed by Web-retrieved resources. The resulting seeds describe scenarios, workflows, and required tools. The harness evaluates how these seeds change the diversity of the accumulated seed pool and uses this feedback to guide strategy creation and revision toward complementary task distributions. The seeds are then instantiated into executable tasks.
To verify task outcomes, we introduce Artifact-Contrastive Rubrics, which compare final artifacts and application states across multiple rollouts to identify consensus, divergence, and shared omissions that inform task-specific acceptance criteria. These criteria are grounded in task requirements and initial facts and applied by an independent judge to retain passing trajectories.
Post-training Qwen3.6-35B-A3B on 4,026 verified trajectories from
tasks generated by 50 evolved strategies yields SeedEvolve-35B-A3B. The model improves Toolathlon-Verified Pass@1 from 36.7\% to 49.1\%,
AutomationBench success rate from 8.5\% to 20.5\%, Job-Bench
score from 21.8 to 34.3, and WorkBuddy Bench (Office) score from 74.8 to 77.8. Further experiments show overall
downstream performance gains as the number of generation
strategies increases. These results highlight the value of
recursive strategy improvement for training working agents.
\end{abstract}
\end{tcolorbox}

\section{Introduction}

\emph{Working agents} carry out practical user requests by retrieving information from the Web, analyzing documents and data, and creating or updating artifacts across applications \cite{xu2024agentcompany,li2025toolathlon}. Completing these workflows requires agents to interpret user goals, gather and reconcile information from multiple sources, plan dependent operations, coordinate tools, and verify the resulting outcomes over long interactions \cite{boisvert2024workarenaplusplus,li2025toolathlon}. Developing these capabilities calls for large-scale, diverse, executable agentic environments and corresponding interaction trajectories that expose models to varied user needs, information dependencies, and workflows \cite{song2026envscaler,dong2026agentworld,mao2026weft,qiao2026ehrcomplex,qiao2026ehrrobustgym}.

\begin{figure}[htbp]
  \centering
  \includegraphics[width=\linewidth]{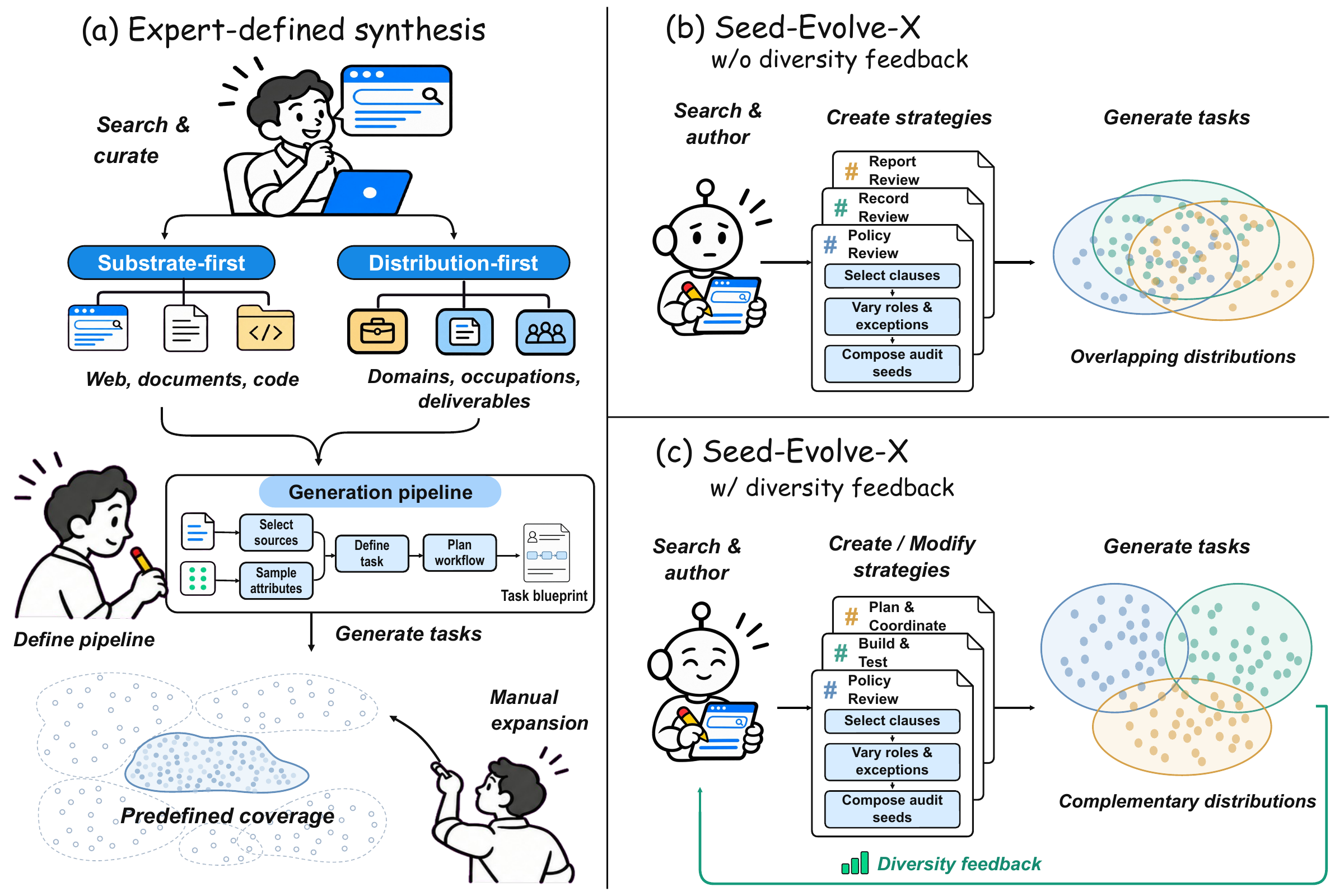}
  \caption{Motivation for feedback-guided strategy evolution.
  (a) Limited coverage under expert-defined synthesis.
  (b) Overlapping distributions from autonomous strategies
  without diversity feedback.
  (c) Complementary distributions through
  diversity-guided strategy evolution.}
  \Description{In panel (a), domain experts collect predefined distributions, including domains, occupations, and deliverable types, together with resources such as skills, code repositories, financial reports, and research papers. An expert-defined synthesis pipeline densely samples within a predefined scope, while further expansion requires manual effort. Panel (b) shows an agent creating policy-review, record-review, and report-review strategies with heavily overlapping task distributions. Panel (c) shows diversity feedback guiding strategy creation and modification toward more complementary directions, illustrated by policy review, building and testing, and planning and coordination.}
  \label{fig:exploration-motivation}
\end{figure}

To obtain such training data, existing synthesis pipelines follow two complementary routes. \emph{Substrate-first} methods derive tasks from selected resources: TaskCraft constructs and composes tool-dependent questions from news webpages, financial and academic PDFs, and embedded charts and tables \cite{shi2025taskcraft}; SWE-smith synthesizes software-repair tasks from code repositories and their tests \cite{yang2025swesmith}; and SkillSynth samples workflow paths from a scenario-mediated skill graph \cite{fan2026skillsynth}. \emph{Distribution-first} methods specify the dimensions used to describe tasks and the candidate values for each dimension, then sample combinations of these values to generate tasks. Examples include Apodex's hierarchy of domains, occupations, and deliverable types \cite{apodex2026apodex11}, ClawGym's combinations of personas, scenario categories, and operations \cite{bai2026clawgym}, and NexForge's research-derived, weighted task profiles and scenario reservoir \cite{zhao2026nexforge}. These routes can also be integrated \cite{zhang2026workforge,zhu2026workgenesis}, with the synthesis pipeline specifying how sampled dimension values are combined with relevant resources to generate individual tasks.

However, these methods still rely heavily on human-defined synthesis pipelines that determine how task dimensions and relevant resources are organized and combined to generate tasks. As illustrated in Figure~\ref{fig:exploration-motivation}(a), continued sampling from a predefined task distribution can limit further gains in task diversity, while broadening that distribution requires additional manual curation or updates to the synthesis pipeline. On the other hand, Figure~\ref{fig:exploration-motivation}(b) illustrates how allowing an agent to autonomously define synthesis strategies risks \emph{mode collapse} \cite{zhang2025verbalizedsampling,lingo2026dynamiccontext}: even distinct strategies may generate highly overlapping task distributions. This highlights the need for diversity feedback in task synthesis: assessing the additional diversity each strategy brings to the existing task collection guides the agent toward complementary task distributions, as illustrated in Figure~\ref{fig:exploration-motivation}(c).

To this end, we propose \textbf{Seed-Evolve-X}, a framework that evolves multiple generation strategies for agentic task synthesis through diversity feedback. An evolution agent searches the Web for real-world resources and uses them to develop seed-generation strategies. Each \emph{strategy} is implemented as a \emph{skill} that defines the dimensions of a task distribution and the candidate values for each dimension. It specifies how to sample these values and combine them with available resources to generate a batch of \emph{task seeds}. Each seed is an extensible blueprint for task instantiation, with three required fields: (1) a \emph{scenario} describing the user's goal and context; (2) a \emph{workflow} describing the operations and dependencies needed to achieve that goal; and (3) the \emph{Model Context Protocol (MCP) services} required to complete the task. Optional extension fields specify additional task details and constraints for instantiation. Seeds provide a common interface for both constructing executable tasks and assessing their diversity relative to the accumulated pool through scenario semantics, workflow semantics, and MCP combinations.

The overall framework consists of two main components. \textbf{(1) Multi-Strategy Evolution Harness.} The evolution harness enables the agent to inspect the current strategy and seed pools, create new strategies, and modify existing ones. For each registered strategy, the harness coordinates seed generation and reports the resulting changes in the accumulated seed pool's scenario diversity, workflow diversity, and MCP-combination coverage. This feedback guides subsequent strategy updates toward complementary task distributions. \textbf{(2) Task Instantiation \& Artifact-Contrastive Rubrics.} Task-specific rubrics are synthesized through retrospective analysis of final artifacts and application states across multiple independent rollouts, rather than inferred solely from task specifications. The rubric author examines \emph{consensus, divergence, and shared omissions} across these outputs, using task requirements and initial facts to formulate acceptance criteria. An independent judge verifies each rollout's final state against the resulting rubric; only passing trajectories are retained for training.

Post-training Qwen3.6-35B-A3B on 4,026 verified trajectories from tasks generated by 50 evolved strategies yields \textbf{SeedEvolve-35B-A3B}. The model improves Toolathlon-Verified~\cite{li2025toolathlon} Pass@1 from 36.7\% to 49.1\%, AutomationBench~\cite{shepard2026automationbench} success rate from 8.5\% to 20.5\%, Job-Bench~\cite{li2026jobbench} score from 21.8 to 34.3, and WorkBuddy Bench (Office)~\cite{tencent2026workbuddy} score from 74.8 to 77.8.

Our contributions can be summarized as follows:
\begin{itemize}
  \item We formulate agentic task synthesis as progressive modeling and sampling of a latent task space, using extensible seeds as a representation for task construction and cross-strategy diversity assessment.
  \item We develop Seed-Evolve-X to autonomously adapt Web-grounded generation strategies through seed-pool diversity feedback, complemented by Artifact-Contrastive Rubrics for final-state verification.
  \item We demonstrate that Seed-Evolve-X generates effective training data for working agents, substantially improving their ability to complete long-horizon work tasks. Further analyses reveal sustained gains in seed-pool diversity across evolution rounds and improved downstream performance as the strategy population expands.
\end{itemize}

\section{Related Work}

\subsection{Agentic Task and Environment Synthesis}
Existing approaches differ in how they select resources and define the tasks to be generated. \textbf{(1) Substrate-first methods} synthesize tasks from existing resources. TaskCraft~\cite{shi2025taskcraft} constructs tool-dependent tasks from webpages and documents, while SWE-smith~\cite{yang2025swesmith} generates repair tasks from repositories and tests. SkillSynth~\cite{fan2026skillsynth} samples workflows from a scenario-mediated skill graph. EnvScaler~\cite{song2026envscaler} derives environment themes from task collections, then constructs environments and tasks.
\textbf{(2) Distribution-first methods} organize generation around task categories. Apodex 1.1~\cite{apodex2026apodex11} uses domains, occupations, and deliverables, while EnvCraft~\cite{zeng2026envcraft} uses scenarios and interaction archetypes to guide environment and task synthesis. \textbf{(3) Hybrid approaches} combine both starting points. ClawGym~\cite{bai2026clawgym} combines persona-conditioned and skill-based synthesis, while AgentSynth~\cite{xie2025agentsynth} uses personas and desktop states to propose subtasks that are executed, verified, and composed into longer tasks.
These methods also adapt. SkillSynth~\cite{fan2026skillsynth} rebalances workflow sampling; TaskCraft~\cite{shi2025taskcraft} optimizes generation prompts. Agent-World~\cite{dong2026agentworld} guides synthesis through capability diagnosis, while WEFT~\cite{mao2026weft} refines interaction systems using execution traces and state evidence.

When adaptation is confined to existing resources and task patterns, additional sampling may still produce similar workflows. Our proposed Seed-Evolve-X therefore evolves the generation strategies themselves, revising resource selection, target scenarios, and workflow construction to expand scenario and workflow coverage.



\suppressfloats[t]
\begin{table}[t]
  \caption{Available MCP service types. Generators receive the complete registered tool schemas, not only service names.}
  \label{tab:available-mcps}
  \centering
  \small
  \begin{tabularx}{\columnwidth}{@{}L{0.46\columnwidth}Y@{}}
    \toprule
    \textbf{MCP service} & \textbf{Capability} \\
    \midrule
    \texttt{terminal} & Commands and scripts \\
    \texttt{filesystem} & File inspection and editing \\
    \texttt{git} & Version and change management \\
    \texttt{excel} & Spreadsheet processing \\
    \texttt{pdf-tools} & Document extraction \\
    \texttt{word} & Text-document editing \\
    \texttt{pptx} & Presentation production \\
    \texttt{playwright\_with\_\allowbreak chunk} & Browser interaction \\
    \texttt{fetch} & Web-content retrieval \\
    \bottomrule
  \end{tabularx}
\end{table}

\begin{table}[!tbp]
  \caption{An example of an agent-generated strategy for Operational Workspace Construction and three corresponding task seeds.}
  \label{tab:operational-workspace-case}
  \centering
  \begingroup
  \small
  \setlength{\tabcolsep}{5pt}
  \renewcommand{\arraystretch}{1.08}
  \definecolor{caseSkillBg}{HTML}{EDF1F6}
  \definecolor{caseSeedBg}{HTML}{FAF1E9}
  \definecolor{caseHeaderBg}{HTML}{EEF3EE}
  \definecolor{caseBlue}{HTML}{385371}
  \definecolor{caseGreen}{HTML}{455D49}
  \definecolor{caseOrange}{HTML}{A65D23}
  \begin{tabularx}{\columnwidth}{YY}
    \toprule
    \multicolumn{2}{l}{\textbf{Operational Workspace Construction}} \\
    \midrule
    \rowcolor{caseHeaderBg}
    \textcolor{caseBlue}{\textbf{Strategy skill ($S$)}} &
    \textcolor{caseOrange}{\textbf{Generated seeds ($s$)}} \\
    \midrule
    \cellcolor{caseSkillBg}
    \textcolor{caseBlue}{\textbf{Summary.}}
    Generate seeds for building or repairing usable operational workspaces.
    \par\smallskip
    \textcolor{caseBlue}{\textbf{Task attributes and variation $D_S$.}}
    Selected attributes and candidate values:
    \par\smallskip
    {\footnotesize
    \textbf{Family:} intake service; field operations; software delivery.
    \par\smallskip
    \textbf{User:} clinic coordinator; field lead; kiosk maintainer.
    \par\smallskip
    \textbf{Materials:} draft forms; custody requirements; repository fixtures.
    \par\smallskip
    \textbf{Output:} routed record; field packet; tested revision.
    \par\smallskip
    \textbf{Dependencies:} service path; field handoff; versioned change.
    \par}
    \par\smallskip
    \textcolor{caseBlue}{\textbf{Web-grounding resources $\mathcal{R}_S$.}}
    \par
    {\footnotesize
    $r_1$: \path{accessible-form-contracts}: staged input and feedback (W3C WAI).
    \par\smallskip
    $r_2$: \path{recoverable-validation-patterns}: visible errors and recovery (GOV.UK).
    \par\smallskip
    $r_3$: \path{change-and-release-evidence}: tested patches and rollback (NIST SSDF).
    \par}
    \par\smallskip
    \textcolor{caseBlue}{\textbf{Seed-generation pipeline $P_S$.}}
    \par
    {\footnotesize
    (1) Select a family and resource-informed construction pattern; vary user, materials, and output.
    \par\smallskip
    (2) Write $c$ and a connected, acyclic $w$ with produced-and-consumed artifacts.
    \par\smallskip
    (3) Specify the system contract, acceptance cases, recovery behavior, and handoff.
    \par\smallskip
    (4) Assign indispensable MCP roles; reject cosmetic variants and redundant tools.
    \par}
    &
    \cellcolor{caseSeedBg}
    \textcolor{caseOrange}{\textbf{1. Referral intake repair}}
    \par
    \textcolor{caseOrange}{\boldmath$c$}: A clinic coordinator needs a referral path repaired.
    \par
    \textcolor{caseOrange}{\boldmath$w$}: Map fields to routing data; repair the browser journey; test submissions; hand off to triage.
    \par
    \textcolor{caseOrange}{\boldmath$\mathcal{M}$}: Filesystem, browser, Excel.
    \par
    \textcolor{caseOrange}{\boldmath$\mathcal{R}$}$=\{r_1,r_2\}$.
    \par
    \textcolor{caseOrange}{\boldmath$\xi$ \textbf{(selected)}}:
    \par
    Artifact handoff: referral interface and routing workbook for triage.
    \par
    Recovery contract: retain inputs and enable correction.
    \par\smallskip
    {\color{black!55}\hrule height 0.35pt}
    \smallskip
    \textcolor{caseOrange}{\textbf{2. Sampling field packet}}
    \par
    \textcolor{caseOrange}{\boldmath$c$}: A field lead needs a stream-visit packet.
    \par
    \textcolor{caseOrange}{\boldmath$w$}: Extract custody requirements; build the packet; test complete and missing-ID handoffs.
    \par
    \textcolor{caseOrange}{\boldmath$\mathcal{M}$}: Filesystem, Word, PDF.
    \par
    \textcolor{caseOrange}{\boldmath$\mathcal{R}$}$=\{r_2\}$.
    \par
    \textcolor{caseOrange}{\boldmath$\xi$ \textbf{(selected)}}:
    \par
    Artifact handoff: sampling packet and handoff format for lab intake.
    \par
    Recovery contract: retain sampling state and site context.
    \par\smallskip
    {\color{black!55}\hrule height 0.35pt}
    \smallskip
    \textcolor{caseOrange}{\textbf{3. Kiosk validation patch}}
    \par
    \textcolor{caseOrange}{\boldmath$c$}: A maintainer needs a validation defect repaired.
    \par
    \textcolor{caseOrange}{\boldmath$w$}: Inspect fixtures; patch; run a repeatable test; hand off the revision with rollback context.
    \par
    \textcolor{caseOrange}{\boldmath$\mathcal{M}$}: Filesystem, Git, terminal.
    \par
    \textcolor{caseOrange}{\boldmath$\mathcal{R}$}$=\{r_3\}$.
    \par
    \textcolor{caseOrange}{\boldmath$\xi$ \textbf{(selected)}}:
    \par
    Artifact handoff: tested revision and rollback reference for the maintainer.
    \par
    Recovery contract: retain prior revision; withhold unverified changes.
    \\
    \bottomrule
  \end{tabularx}
  \endgroup
\end{table}


\subsection{Recursive Self-Improvement}
Self-improvement has been studied at several levels, including model parameters~\cite{huang2025rzero}, prompts~\cite{agrawal2025gepa,fernando2024promptbreeder}, contexts~\cite{zhang2025ace}, skills~\cite{zhang2026coevoskills}, execution harnesses~\cite{lee2026metaharness}, and environments~\cite{huang2026envharness}. These methods differ in both the components they optimize and the feedback they use. For example, Meta-Harness~\cite{lee2026metaharness} uses prior candidates' source code, evaluation scores, and execution traces to propose changes to harness code. RSIBench-Data~\cite{meng2026rsibench} evaluates agents that synthesize training data, train a target model, and use the resulting evaluation feedback to revise their data-generation strategies.

Seed-Evolve-X evolves task-generation strategies using the diversity of task seeds as a proxy for the diversity of synthesized tasks. Feedback is computed directly from changes in the accumulated seed pool, bypassing the longer feedback loop of task instantiation and execution. This reduces the cost of strategy evaluation, enabling broader exploration under a given computational budget.


\section{Method}

\begin{figure*}[t]
  \centering
  \includegraphics[width=\textwidth,trim=19.9bp 75.2bp 247.5bp 12.1bp,clip]{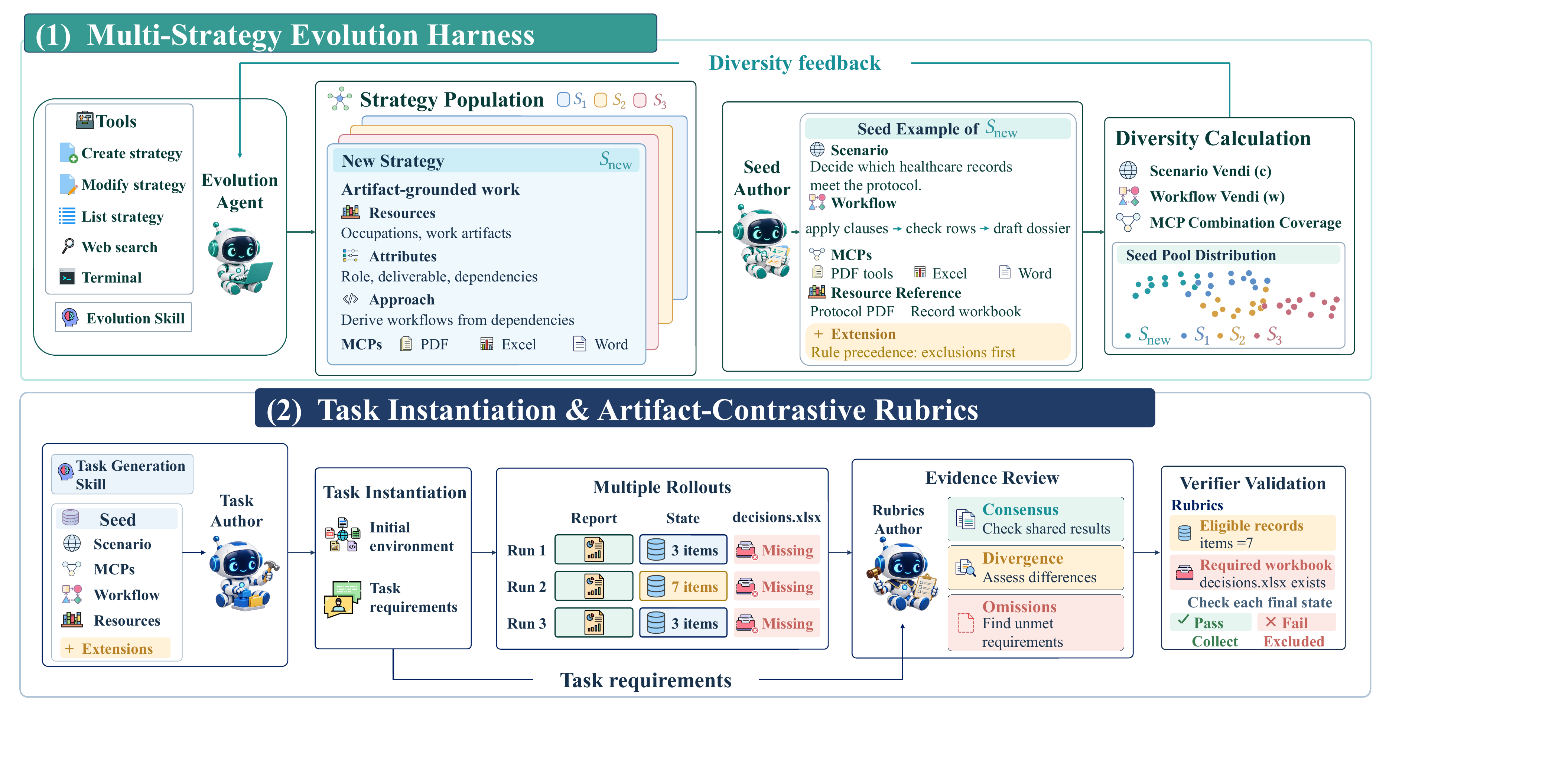}
  \caption{Seed-Evolve-X overview. (1) The Multi-Strategy Evolution Harness supports Web-grounded strategy creation and modification, seed generation, and feedback on scenario diversity, workflow diversity, and MCP-combination coverage. (2) Seeds and their referenced resources are instantiated into executable tasks. Artifact-Contrastive Rubrics examine consensus, divergence, and shared omissions across rollout artifacts against task requirements and initial facts. An independent judge verifies each rollout's final state; only passing trajectories enter the training pool.}
  \Description{Two connected modules. The upper module creates and modifies strategies, generates seeds with scenarios, workflows, MCP choices and extensions, and returns diversity feedback. A selected seed and its resources feed task instantiation in the lower module. Multiple rollouts produce final artifacts and states. A rubric author examines shared results, differing record counts and missing outputs to define final-state acceptance criteria. Validation separates passing from failing trajectories; the passing branch is collected for training, while the failing branch is excluded.}
  \label{fig:seed-evolve-overview}
\end{figure*}

Seed-Evolve-X is a framework for agentic task synthesis that autonomously discovers real-world resources, creates multiple generation strategies, and evolves them through diversity feedback. As shown in Figure~\ref{fig:seed-evolve-overview}, the framework comprises two components. \textbf{(1) Multi-Strategy Evolution Harness.} The harness coordinates strategy creation and modification, seed generation, and diversity assessment of the accumulated seed pool to guide exploration toward complementary task distributions. \textbf{(2) Task Instantiation \& Artifact-Contrastive Rubrics.} This component instantiates seeds into executable tasks and constructs task-specific rubrics by comparing final artifacts across multiple rollouts. Only trajectories that pass verification are retained for training.

\subsection{Generation Strategies and Task Seeds}
\label{sec:seed-representation}

We first define the generation strategies that the evolution
harness creates and revises and the task seeds used for diversity
assessment and task instantiation.

\noindentparagraph{\textbf{\upshape Generation strategies.}}
Let $\mathcal{T}^{\star}$ denote the space of agentic tasks.
We view each strategy $S$ as defining a target region
$\mathcal{T}_S \subseteq \mathcal{T}^{\star}$ in terms of its task
attributes, resources, and generation rules.
Strategy evolution changes which regions are explored and how
tasks are generated within them.
Each strategy is implemented as a reusable skill:
\begin{equation}
  S=(D_S,\mathcal{E}_S,\mathcal{R}_S,P_S).
\end{equation}
Here, $D_S$ specifies task attributes, their candidate values,
and allowed variation;
$\mathcal{E}_S$ defines strategy-specific extension fields and
their constraints;
$\mathcal{R}_S$ contains relevant resources;
and $P_S$ specifies construction rules, batch-level variation
requirements, and quality checks for generating seeds.

During strategy authoring, the evolution agent can retrieve
Web resources and record their sources in $\mathcal{R}_S$.
These resources can supply documents, code, or data for initial
task environments, or inform task attributes and generation
rules through descriptions of occupations, workflows, and
professional requirements.
A strategy can combine both uses, and $\mathcal{R}_S$ may be
empty when no external resources are needed.

\noindentparagraph{\textbf{\upshape Task seeds.}}
A task seed is an extensible blueprint for task instantiation:
\begin{equation}
  s=(c,w,\mathcal{M},\mathcal{R},\xi).
\end{equation}
The three required base fields are the \emph{scenario} $c$,
describing the user's role, context, and goal;
the \emph{workflow} $w$, describing the operations and information
dependencies needed to achieve that goal; and the required
\emph{MCP services} $\mathcal{M}$.
Table~\ref{tab:available-mcps} lists the nine available service types.
The resource field $\mathcal{R}$ references resources in
$\mathcal{R}_S$ and may be empty.
Optional extension fields capture additional task details,
such as rule precedence, deliverable requirements, or workflow
constraints. When specified by the strategy, their values
$\xi=(e_1,\dots,e_n)$ must follow the schema and constraints
defined in $\mathcal{E}_S$.

The evolution harness uses scenario and workflow embeddings
as proxies for semantic similarity across seeds and separately
measures MCP-combination coverage.
The task author uses the seed fields and referenced resources
to construct an executable task.
Seeds thus provide a common interface between strategy evolution
and task instantiation.

\noindentparagraph{\textbf{\upshape Illustrative example.}}
Table~\ref{tab:operational-workspace-case} shows an operational
workspace construction strategy and three generated seeds.
The strategy combines task attributes, resource-informed
construction rules, and a shared extension schema to generate
tasks involving referral intake, field sampling, and kiosk repair.
For example, guidance from W3C WAI form tutorials informs
the staged input and correction feedback in the referral-intake
seed.
The seeds specify different workflows, MCP combinations,
deliverables, and recovery conditions, illustrating how one
strategy supports varied tasks under shared generation rules.

\subsection{Multi-Strategy Evolution Harness}
\label{sec:strategy-evolution}

The Multi-Strategy Evolution Harness guides strategy creation and
revision to expand scenario and workflow coverage beyond the tasks
already represented in the seed pool. It connects strategy proposal,
independent seed generation, and diversity assessment in a feedback
loop, using the generated seeds to inform subsequent strategy updates.

\noindentparagraph{\textbf{\upshape Strategy proposal and seed generation.}}
At iteration $t$, active strategies form population $\mathcal{S}_t$,
and their seed batches form pool $\mathcal{B}_t$.
The evolution agent receives the seed schema, quality requirements,
and complete MCP tool schemas. It inspects the current population,
generated seeds, and historical feedback reports to choose an
exploration direction, gather resources, and author a candidate skill.
Registration binds the
candidate instructions to a resource pack and records its version.
The harness supports two operations:
\begin{enumerate}[leftmargin=*,nosep]
  \item \textbf{Create}: register a new independent strategy and add
  its generated seed batch to the pool.
  \item \textbf{Modify}: propose a replacement for an existing strategy
  and evaluate its seeds before accepting or discarding the replacement.
\end{enumerate}
Creation expands the strategy population; modification revises an
existing member without adding another active strategy.

For each registered candidate, the harness launches an independent
seed-generation agent with the strategy instructions, resource pack,
and available tool interfaces. The generator produces $N$ seeds,
each containing the required base fields, strategy-defined extensions,
and resource references. The evolution agent can inspect individual
seeds to relate aggregate diversity measurements to the scenarios,
workflows, and tool combinations actually generated.

\noindentparagraph{\textbf{\upshape Diversity assessment.}}
The harness evaluates each generated batch and the accumulated seed
pool along three axes: scenario semantics, workflow semantics, and
MCP combinations. It reports these measurements and their changes
relative to the current accepted pool, allowing the agent to assess
each candidate's contribution to the existing collection.

For a pool $\mathcal{B}=\{s_i\}_{i=1}^{n}$, we encode scenarios $c_i$
and workflows $w_i$ separately and compute their Vendi
scores~\cite{friedman2023vendi}. For $x\in\{c,w\}$, an embedding
model $f$ gives
\begin{equation}
  z_i^{(x)}=\frac{f(x_i)}{\|f(x_i)\|_2},\qquad
  K_{ij}^{(x)}=\langle z_i^{(x)},z_j^{(x)}\rangle.
\end{equation}
Let $\lambda_j^{(x)}$ be the nonzero eigenvalues of $K^{(x)}/n$.
Then
\begin{equation}
  D_x(\mathcal{B})=\exp\left(-\sum_j
  \lambda_j^{(x)}\log\lambda_j^{(x)}\right).
\end{equation}
$D_c$ and $D_w$ measure effective diversity in their respective
embedding spaces. Computing them separately distinguishes variation
in user goals and contexts from variation in the operations and
dependencies needed to complete the tasks.

For MCP-combination coverage, we treat each seed's MCP configuration
as an unordered set. The set of distinct MCP combinations in the pool is
\begin{equation}
  \mathcal{C}(\mathcal{B})=\{\mathcal{M}_i:s_i\in\mathcal{B}\}.
\end{equation}
We compute coverage as the size-weighted count of these combinations:
\begin{equation}
  D_{\mathrm{MCP}}(\mathcal{B})=\sum_{A\in\mathcal{C}(\mathcal{B})}|A|.
\end{equation}
This equals the number of unique combinations times their mean size.
Repeating an existing combination adds no coverage. To keep the
measure tied to task requirements, generation instructions prohibit
adding irrelevant services merely to increase coverage.

\noindentparagraph{\textbf{\upshape Feedback-guided strategy updates.}}
Feedback reports signed differences between the candidate's projected
pool and the current accepted pool; these differences may be negative.
The agent uses these changes together with generated seeds and
nearest-neighbor examples to diagnose recurring scenarios and workflow
patterns and inform subsequent strategy creation or revision.
For a modification, the projected pool replaces the old strategy's
batch with the candidate batch, while the old version and its seeds
remain active during evaluation. The agent then accepts the
replacement or retains the original. Accepted changes update the
strategy population, seed pool, and report used for the next decision.
This loop continues within the strategy and modification budgets,
guiding exploration toward complementary task distributions without
assuming that every update improves every diversity measure.
Appendix~\ref{sec:strategy-case-study} illustrates this process
through a recorded case of feedback-guided strategy revision.

\subsection{Task Instantiation \& Artifact-Contrastive Rubrics}

This component turns task seeds into executable tasks and selects
verified trajectories for training. For tasks involving multiple
artifacts or application-state changes, verification requires concrete
checks of the requested outcomes and acceptable variation.
We construct task-specific rubrics using evidence from multiple
independent executions, then apply them to verify each rollout's
final state.

\noindentparagraph{\textbf{\upshape Task instantiation and independent rollouts.}}
Given seed $s_i$ and its referenced resources $\mathcal{R}_i$,
the task author produces
\begin{equation}
  (u_i,\mathcal{E}_i^0,\mathcal{A}_i)
  =\mathrm{Author}(s_i,\mathcal{R}_i).
\end{equation}
Here, $u_i$ is the user request, $\mathcal{E}_i^0$ contains the
initial files and service data, and $\mathcal{A}_i$ specifies the
available tools. The request expresses the desired outcome in a
natural user voice, while the initial environment supplies the
materials and facts needed to act. Instantiation preserves the
seed's workflow dependencies, required tools, and extension constraints.

Each task is executed independently $K$ times from the same initial
environment. Run $k$ produces a trajectory $\tau_{i,k}$ and final
state $\mathcal{E}_{i,k}^{\mathrm{end}}$. The trajectory records
messages, tool calls, and observations; the final state contains
generated files, updated records, and observable application states.
We collect these execution records as
\begin{equation}
  \mathcal{O}_i=\{(\tau_{i,k},\mathcal{E}_{i,k}^{\mathrm{end}})\}_{k=1}^{K}.
\end{equation}

\noindentparagraph{\textbf{\upshape Artifact-contrastive rubric synthesis.}}
The rubric author compares final artifacts and application states
across rollouts, examining \emph{consensus, divergence, and shared
omissions}. Consensus identifies recurring outcomes; divergence
reveals differences in artifact contents or application states;
shared omissions identify required outputs or state changes absent
across all runs. Execution records can be consulted to clarify
disagreements or apparent omissions. These observations help identify
missing checks and refine acceptance criteria, while task requirements
and initial facts determine correctness. Agreement across executions
alone does not establish that an outcome is correct.

The author constructs a task-specific rubric:
\begin{equation}
  \mathcal{V}_i=\mathrm{RubricAuthor}(u_i,\mathcal{E}_i^0,\mathcal{O}_i).
\end{equation}
Each item $(v_{i,j},a_{i,j})$ pairs an acceptance condition with its
assigned points. Conditions specify required values, artifact contents,
service states, and acceptable variation. They evaluate observable
outcomes and allow alternative execution paths that satisfy the task.

\begin{table*}[!t]
  \definecolor{mainTableGain}{HTML}{D35400}
  \newcommand{\scoreGain}[2][\uparrow]{\makebox[0pt][l]{\kern0.3pt\textsubscript{\textcolor{mainTableGain}{\ensuremath{#1}#2}}}}
  \centering
  \caption{Downstream performance on Toolathlon-Verified, AutomationBench, Job-Bench, and WorkBuddy Bench (Office). Rates are percentages; scores use a 0--100 scale. Superscript $\dagger$ denotes results reported by external sources; unmarked results are reproduced by us. Colored subscripts give absolute changes from Qwen3.6-35B-A3B, computed before rounding. Dashes indicate unavailable or unverified measurements. Benchmark details and evaluation protocols are provided in Appendix~\ref{sec:evaluation-score-details}.}
  \label{tab:main-results}
  \small
  \setlength{\tabcolsep}{6pt}
  \renewcommand{\arraystretch}{1.0}
  \definecolor{mainTableGroup}{RGB}{232,238,247}
  \resizebox{\linewidth}{!}{%
  \begin{tabular}{l*{7}{c}}
    \toprule
     & \multicolumn{3}{c}{\textbf{Toolathlon-Verified}} & \multicolumn{2}{c}{\textbf{AutomationBench}} & \textbf{Job-Bench} & \shortstack{\textbf{WorkBuddy Bench}\\\textbf{(Office)}} \\
    \cmidrule(lr){2-4}\cmidrule(lr){5-6}\cmidrule(lr){7-7}\cmidrule(l){8-8}
    \textbf{Model} & Pass@1 $\uparrow$ & Pass@3 $\uparrow$ & Pass\textasciicircum{}3 $\uparrow$ & Avg. score $\uparrow$ & Success rate $\uparrow$ & Score $\uparrow$ & Score $\uparrow$ \\
    \midrule
    \rowcolor{mainTableGroup}
    \multicolumn{8}{c}{\textit{Proprietary Models}} \\
    Claude Opus 4.8 (max) & \textbf{76.2}$^{\dagger}$ & \textbf{84.3}$^{\dagger}$ & \textbf{66.7}$^{\dagger}$ & -- & \textbf{48.0}$^{\dagger}$ & \textbf{48.4}$^{\dagger}$ & \textbf{81.8} \\
    GPT-5.6 Sol (xhigh) & 74.9$^{\dagger}$ & -- & -- & -- & 45.8$^{\dagger}$ & 45.4$^{\dagger}$ & 80.8 \\
    \midrule
    \rowcolor{mainTableGroup}
    \multicolumn{8}{c}{\textit{Open-Weight Models}} \\
    DeepSeek-V4-Flash-0731 & \textbf{70.7}$^{\dagger}$ & \textbf{81.5}$^{\dagger}$ & \textbf{58.3}$^{\dagger}$ & -- & 22.2 & 41.3$^{\dagger}$ & \textbf{80.9} \\
    GLM-5.2 & 59.9$^{\dagger}$ & 78.7$^{\dagger}$ & 41.7$^{\dagger}$ & -- & \textbf{26.2}$^{\dagger}$ & \textbf{43.4}$^{\dagger}$ & 79.6$^{\dagger}$ \\
    Qwen3.5-397B-A17B & 40.7$^{\dagger}$ & 51.9$^{\dagger}$ & 27.8$^{\dagger}$ & -- & 4.2$^{\dagger}$ & 19.1 & 73.1 \\
    Qwen3.6-27B & 34.3 & 48.1 & 20.4 & -- & 9.7$^{\dagger}$ & 21.8$^{\dagger}$ & 65.7$^{\dagger}$ \\
    \midrule
    \rowcolor{mainTableGroup}
    \multicolumn{8}{c}{\textit{Open-Weight Models (35BA3B parameters)}} \\
    Qwen3.6-35B-A3B & 36.7 & 46.3 & 22.2 & 40.6 & 8.5 & 21.8 & 74.8 \\
    Fx-Work-35B & 38.3\scoreGain{1.6} & 51.9\scoreGain{5.6} & 22.2 & 52.7\scoreGain{12.1} & 16.7\scoreGain{8.2} & 25.2\scoreGain{3.4} & 76.3\scoreGain{1.5} \\
    \rowcolor{black!4}
    \textbf{SeedEvolve-35B-A3B} & \textbf{49.1}\scoreGain{12.4} & \textbf{64.8}\scoreGain{18.5} & \textbf{34.3}\scoreGain{12.0} & \textbf{58.0}\scoreGain{17.4} & \textbf{20.5}\scoreGain{12.0} & \textbf{34.3}\scoreGain{12.4} & \textbf{77.8}\scoreGain{3.0} \\
    \bottomrule
  \end{tabular}
  }
\end{table*}

\noindentparagraph{\textbf{\upshape Final-state verification.}}
The rubric is fixed before judging and reused across the task's
rollouts. An independent judge inspects each rollout's actual final
artifacts and application states using read-only tools, without
accessing its execution history. Let $a_{i,j}$ denote the points
assigned to rubric item $j$ for task $i$, and let
$b_{i,k,j}\in\{0,1\}$ indicate whether rollout $k$ fully satisfies
that item. The total verification score $\rho_{i,k}$ is
\begin{equation}
  \rho_{i,k}=\sum_{j=1}^{|\mathcal{V}_i|}a_{i,j}b_{i,k,j}.
\end{equation}
Each item receives its assigned points only when fully satisfied,
making scores comparable across rollouts of the same task.
The evaluator applies its configured pass threshold together with
any task-specific deterministic checks, such as required-file checks.
Only valid trajectories that pass verification enter the training pool.
The rubric-author and judge prompts, together with their input and
output formats, are provided in Appendix~\ref{sec:rubric-implementation}.

\section{Experiments}

\subsection{Experimental Setup}
\noindentparagraph{\textbf{\upshape Models and Training Data.}}
We fine-tune Qwen3.6-35B-A3B on verified trajectories generated by DeepSeek-V4-Flash-0731. Strategy evolution allows up to two modifications per strategy and produces a final pool of 2,500 seeds from 50 retained strategies, with 50 seeds per strategy. Each seed is instantiated into one executable task, and the teacher performs three independent rollouts per task, producing 7,500 candidate trajectories. Rubric-based verification retains 4,026 passing trajectories for supervised fine-tuning (SFT), yielding \textbf{SeedEvolve-35B-A3B}.

\noindentparagraph{\textbf{\upshape Baselines.}}
We compare SeedEvolve-35B-A3B with its starting checkpoint, Qwen3.6-35B-A3B, to assess the gains from training on our synthesized data. We also include Fx-Work-35B, initialized from the same backbone, as a reference for agent post-training. Fx-Work-35B is fine-tuned on occupational tasks constructed from real-world artifacts through WorkGenesis's evidence-based construction and execution-guided verification pipeline \cite{zhu2026workgenesis}. For broader capability comparisons, we include proprietary models (Claude Opus 4.8 with max reasoning effort and GPT-5.6 Sol with xhigh reasoning effort) and other open-weight models: DeepSeek-V4-Flash-0731, GLM-5.2, Qwen3.5-397B-A17B, and Qwen3.6-27B.

\noindentparagraph{\textbf{\upshape Benchmarks.}}
We evaluate on four benchmarks covering complementary working-agent capabilities: Toolathlon-Verified for long-horizon tool use across applications \cite{li2025toolathlon,toolathlon2026leaderboard}, AutomationBench for business-workflow execution \cite{shepard2026automationbench}, Job-Bench for professional deliverable production \cite{li2026jobbench}, and WorkBuddy Bench (Office) for office workflows involving mixed-format files and workspace operations \cite{tencent2026workbuddy}. We report Pass@1, Pass@3, and Pass$^3$ on Toolathlon-Verified; average score and success rate on AutomationBench; and aggregate scores on Job-Bench and WorkBuddy Bench (Office). Benchmark configurations, metric definitions, and evaluation protocols are detailed in Appendix~\ref{sec:evaluation-score-details}.

\noindentparagraph{\textbf{\upshape Implementation Details.}}
We use GPT-5.6-Terra for the strategy-evolution agent, seed-generation agents, task-instantiation author, rubric author, and final-state judge. Qwen3-Embedding-8B encodes scenarios and workflows separately for semantic diversity assessment. We perform SFT using the Megatron backend of ms-swift for five epochs, with a maximum sequence length of 262k tokens, sequence packing, and a global batch size of 16. The learning rate is $10^{-5}$, with a warmup fraction of 5\% and a minimum learning rate of $10^{-6}$. The language-model parameters are fine-tuned while the vision encoder and multimodal aligner remain frozen.

\subsection{Main Results}
\noindentparagraph{\textbf{\upshape Comparison with Same-Backbone Models.}}
Table~\ref{tab:main-results} shows that SeedEvolve-35B-A3B consistently improves over its Qwen3.6-35B-A3B starting checkpoint across all four benchmarks. SFT on 4,026 verified trajectories from 50 evolved strategies improves Toolathlon-Verified Pass@1 by 12.4 percentage points, AutomationBench success rate by 12.0 percentage points, Job-Bench score by 12.4 points, and WorkBuddy Bench (Office) score by 3.0 points. On AutomationBench, the average score also increases from 40.6 to 58.0, indicating that the gains extend to both graded task performance and complete-task success. Beyond the original checkpoint, we compare with Fx-Work-35B, which uses the same backbone and a predefined pipeline for occupational task construction \cite{zhu2026workgenesis}. SeedEvolve-35B-A3B outperforms Fx-Work-35B by 10.8 percentage points in Toolathlon-Verified Pass@1, 3.8 percentage points in AutomationBench success rate, and 1.5 points on WorkBuddy Bench (Office). These results demonstrate that training data synthesized through autonomously evolved strategies can support competitive agent post-training without relying on a manually specified generation pipeline.

\noindentparagraph{\textbf{\upshape Comparison with Larger and Proprietary Models.}}
SeedEvolve-35B-A3B achieves competitive results on selected benchmarks relative to larger open-weight models. On Toolathlon-Verified, it exceeds Qwen3.5-397B-A17B by 8.4 percentage points (49.1\% versus 40.7\%) despite having substantially fewer total parameters. On WorkBuddy Office, it scores 77.8, approaching GLM-5.2's 79.6. Its AutomationBench success rate remains below that of its demonstration teacher, DeepSeek-V4-Flash-0731 (20.5\% versus 22.2\%). Substantial gaps to proprietary models remain: Claude Opus 4.8 achieves 76.2\% on Toolathlon-Verified and 48.4 on Job-Bench, compared with our model's 49.1\% and 34.3. These comparisons indicate that the synthesized data supports competitive performance in specific working-agent settings, while broad frontier-level capability remains out of reach.

\Needspace{8\baselineskip}
\subsection{Ablation Studies}
\label{sec:feedback-blind-analysis}
\begin{wrapfigure}{r}{0.49\textwidth}
  \vspace{-\intextsep}
  \centering
  \includegraphics[width=\linewidth]{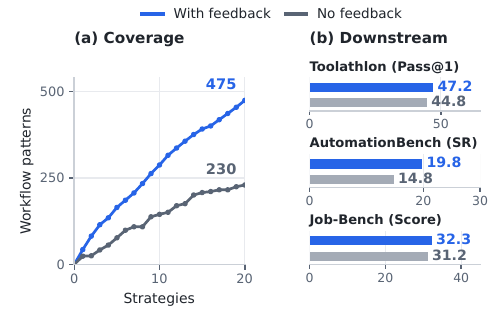}
  \caption{Effects of diversity feedback at a 20-strategy budget. (a) Cumulative execution-pattern coverage. (b) Downstream performance.}
  \Description{The left panel plots cumulative distinct workflow clusters against the number of strategies, from zero to 20 in each condition. It uses all 955 feedback and 733 no-feedback completed task traces, with tool-function names only and a shared complete-linkage clustering at normalized LCS similarity 0.5. Coverage at 20 strategies is 475 versus 230 patterns. The right panel compares feedback and no-feedback performance: Toolathlon-Verified Pass@1 47.2 versus 44.8 percent, AutomationBench success rate 19.8 versus 14.8 percent, and Job-Bench score 32.3 versus 31.2.}
  \label{fig:feedback-blind-analysis}
\end{wrapfigure}

\textbf{\upshape Diversity Feedback in Evolution.}\space
We examine whether diversity feedback broadens the execution patterns of synthesized tasks and improves downstream agent performance. We compare evolution runs with and without feedback at a budget of 20 strategies, each generating 50 seeds, and fine-tune the same Qwen3.6-35B-A3B checkpoint on the corresponding verified trajectories.

Without feedback, workflow Vendi decreases by 64.5\% and MCP-combination coverage falls from 754 to 76, compared with a 7.7\% decrease in scenario Vendi. To compare executions, we retain function names only and collapse consecutive identical calls. For nonempty sequences $A,B$, normalized longest common subsequence (LCS) similarity is
\[
  s(A,B)=\frac{2\lvert\operatorname{LCS}(A,B)\rvert}{\lvert A\rvert+\lvert B\rvert}.
\]
We pool both conditions and apply complete-linkage clustering with within-cluster similarity at least 0.5. Coverage counts distinct clusters represented by tasks from the first $k$ strategies. Figure~\ref{fig:feedback-blind-analysis}(a) uses one normally completed trace per task\footnote{We select the lowest-index completed original rollout, falling back to archived retries, without filtering by rubric score. Historical runs differ in strategy modification/selection and retry availability.} and shows 475 patterns with feedback versus 230 without feedback at 20 strategies, based on 955 and 733 tasks, respectively. At an equal budget of 733 completed tasks, coverage remains higher with feedback: 396 versus 230 patterns.

Figure~\ref{fig:feedback-blind-analysis}(b) shows that training on feedback-guided data also yields stronger downstream performance: Toolathlon-Verified Pass@1 is 47.2\% versus 44.8\%, AutomationBench success rate is 19.8\% versus 14.8\%, and the Job-Bench score is 32.3 versus 31.2. Together, these results support the benefits of diversity-guided exploration beyond seed descriptions, with broader observed execution coverage and stronger performance from the trained agent.

\noindentparagraph{\textbf{\upshape Verification Mechanism.}}
\label{sec:verification-ablation}
To assess the value of trajectory verification and artifact-based rubric construction, we compare three conditions: (i) \emph{no verification}, which applies no task-completion filtering; (ii) \emph{requirements-only rubrics}, derived from task requirements and the initial environment; and (iii) \emph{Artifact-Contrastive Rubrics}, which additionally use consensus, divergence, and shared omissions across rollout artifacts and application states to formulate verification criteria. Of the 7,500 candidate trajectories, the no-verification condition retains 6,288 after excluding runs with execution errors and those that reach the maximum number of turns without finishing. Requirements-only rubrics retain 3,402 trajectories, while Artifact-Contrastive Rubrics retain 4,026. We fine-tune on the trajectories retained by each condition and report downstream performance alongside training-set sizes in Table~\ref{tab:verification-ablation}.

\Needspace{14\baselineskip}
\begin{wraptable}{r}{0.55\textwidth}
  \centering
  \small
  \setlength{\tabcolsep}{5pt}
  \renewcommand{\arraystretch}{1.1}
  \caption{Verification mechanisms, SFT data size, and downstream performance. T: Toolathlon Pass@1; A: AutomationBench success rate; J: Job-Bench score.}
  \label{tab:verification-ablation}
  \begin{tabular}{@{}lrrrr@{}}
    \toprule
    \textbf{Verification} & \shortstack{\textbf{SFT}\\\textbf{trajectories}} & \textbf{T} & \textbf{A} & \textbf{J} \\
    \midrule
    No verification & 6,288 & 45.7 & 18.0 & 30.4 \\
    Requirements-only Rubrics & 3,402 & 46.3 & 19.7 & 32.4 \\
    Artifact-Contrastive Rubrics & 4,026 & \textbf{49.1} & \textbf{20.5} & \textbf{34.3} \\
    \bottomrule
  \end{tabular}
\end{wraptable}

Despite retaining the largest training set (6,288 trajectories), no verification yields the lowest performance on all three benchmarks. Requirements-only rubrics achieve better results with only 3,402 trajectories, supporting the value of task-completion verification over simply retaining more training data. Beyond this benefit, Artifact-Contrastive Rubrics further outperform requirements-only rubrics, improving Toolathlon-Verified Pass@1 from 46.3\% to 49.1\%, AutomationBench success rate from 19.7\% to 20.5\%, and Job-Bench score from 32.4 to 34.3. These additional gains support incorporating cross-rollout artifact comparisons into rubric construction for training-data selection.

\Needspace{32\baselineskip}
\subsection{Analysis of Strategy Evolution}

\begin{wrapfigure}{r}{0.49\textwidth}
  \vspace{-\intextsep}
  \centering
  \includegraphics[width=\linewidth,trim=0 0 0 22bp,clip]{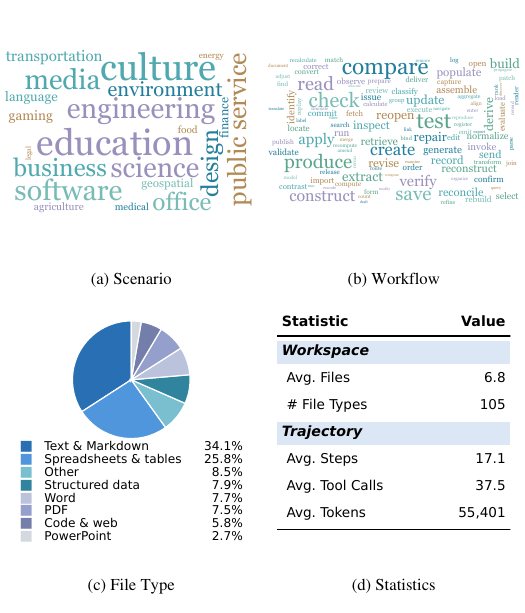}
  \Description{Four panels in a two-by-two grid. Top left: a, scenario domains in 2,500 seeds. Top right: b, declared workflow actions in the same seed pool. Bottom left: c, initial-workspace file-type proportions over 11,756 files from the 1,735 distinct tasks represented in the training data. Bottom right: d, a statistics table showing 6.8 files per initial workspace, 105 file extensions, and averages of 17.1 assistant messages, 37.5 tool calls, and 55,401 tokens per complete serialized trajectory across 4,026 retained training trajectories.}
  \caption{Characteristics of the seed pool generated by 50 evolved strategies and the resulting training data. (a) Scenario word cloud. (b) Workflow word cloud. (c) Initial-workspace file types. (d) Workspace and trajectory statistics.}
  \label{fig:task-distribution}
\end{wrapfigure}

\textbf{\upshape Dataset Characteristics.}\space
Figure~\ref{fig:task-distribution} characterizes the synthesized data along two complementary dimensions: the contexts in which agents work and the operations they are expected to perform. Across 2,500 seeds from 50 evolved strategies, panel (a) spans domains ranging from culture and education to engineering and software, showing that task generation extends beyond a single occupational setting. Panel (b) reveals a mix of information inspection and validation (e.g., compare, check, and test) and artifact construction and modification (e.g., create, produce, and update). Together, these patterns suggest that the generated task pool varies not only in subject matter but also in the kinds of work required from an agent.

Panels (c) and (d) highlight the richness of initial task materials and the long-horizon nature of task execution. The 1,735 distinct tasks represented in the 4,026 retained trajectories contain 11,756 initial-workspace files spanning 105 file extensions, with an average of 6.8 files per task. Text and Markdown files account for 34.1\% of these materials, while spreadsheets and tables account for 25.8\%, complemented by Word documents, PDFs, presentations, structured data, and code. This breadth exposes agents to heterogeneous information representations and artifact formats across the task pool. Trajectories average 17.1 assistant steps, 37.5 tool calls, and 55,401 tokens, reflecting the long-horizon nature of the synthesized tasks. Together, these characteristics highlight the framework's ability to synthesize complex, long-horizon tasks grounded in heterogeneous working materials for training working agents.

\begin{figure}[!htbp]
  \centering
  \includegraphics[width=\columnwidth]{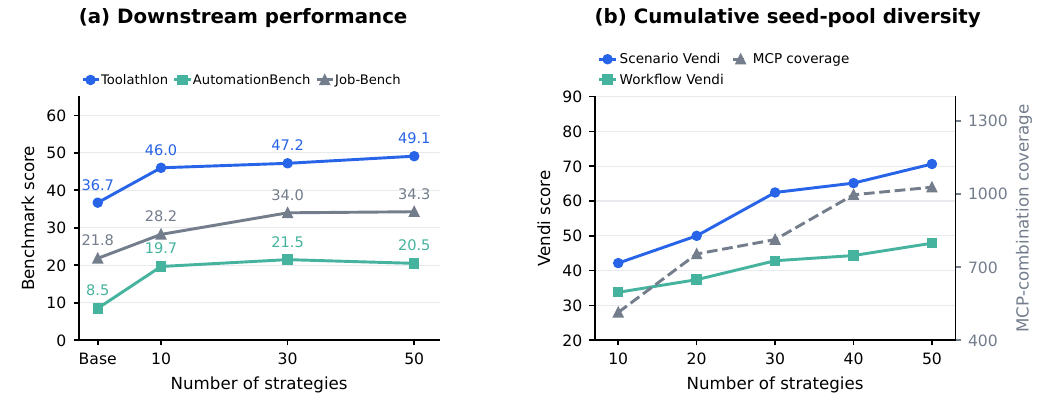}
  \caption{Scaling with the number of strategies. (a) Toolathlon Pass@1 (\%), AutomationBench success rate (\%), and Job-Bench score. (b) Cumulative seed-pool scenario and workflow Vendi (left axis) and MCP-combination coverage (right axis), with 50 seeds per strategy.}
  \Description{Two side-by-side line charts. Panel (a) compares three benchmarks at Base, 10, 30, and 50 strategies: Toolathlon Pass@1 is 36.7, 46.0, 47.2, and 49.1; AutomationBench success rate is 8.5, 19.7, 21.5, and 20.5; Job-Bench scores are 21.8, 28.2, 34.0, and 34.3. Panel (b) shows cumulative seed-pool diversity at 10, 20, 30, 40, and 50 strategies. Scenario Vendi is 42.2, 50.0, 62.4, 65.1, and 70.6; workflow Vendi is 33.8, 37.4, 42.8, 44.3, and 47.9 on the left axis. MCP-combination coverage is 514, 754, 813, 997, and 1,028 on the right axis.}
  \label{fig:strategy-scaling}
\end{figure}

\noindentparagraph{\textbf{\upshape Scaling with the Number of Strategies.}}
To examine how continued strategy exploration affects task diversity and downstream performance, we expand the strategy population to 50, with each strategy contributing 50 seeds, and fine-tune Qwen3.6-35B-A3B on data from 10, 30, and 50 strategies under the same training configuration. Figure~\ref{fig:strategy-scaling}(a) shows overall performance gains as the strategy set expands. From 10 to 50 strategies, Toolathlon Pass@1 increases from 46.0\% to 49.1\%, and the Job-Bench score rises from 28.2 to 34.3. AutomationBench success rate improves from 19.7\% to 20.5\%, although it peaks at 21.5\% with 30 strategies. The different trends may reflect differences between workspace-oriented tasks in Toolathlon and Job-Bench and simulated SaaS workflows in AutomationBench.

\FloatBarrier

Figure~\ref{fig:strategy-scaling}(b) shows that the growing seed pool also covers a broader range of task semantics and service combinations. From 10 to 50 strategies, scenario Vendi increases from 42.2 to 70.6, workflow Vendi from 33.8 to 47.9, and MCP-combination coverage from 514 to 1,028. These increases indicate that continued exploration expands the diversity of the accumulated pool alongside its size. Together, the results support scaling task synthesis through continued strategy exploration, with downstream gains that vary across different types of evaluation tasks.

\FloatBarrier
\section{Conclusion}
We presented Seed-Evolve-X, an automated framework for agentic task synthesis that discovers real-world resources and evolves complementary generation strategies through seed-pool diversity feedback. Extensible task seeds connect strategy exploration to executable-task construction, while Artifact-Contrastive Rubrics support trajectory verification through comparisons of rollout artifacts and application states. Post-training on 4,026 verified trajectories improves Qwen3.6-35B-A3B across four benchmarks. Ablation studies support the contributions of diversity feedback and artifact-based verification, while scaling analyses show growth in cumulative seed-pool diversity and overall downstream gains as the strategy population expands. These findings establish autonomously evolved generation strategies as a practical approach to producing diverse, verifiable training tasks for working agents.

\bibliographystyle{colm2024_conference}
\bibliography{references}
\appendix
\begin{table*}[t]
  \centering
  \caption{A recorded feedback-guided strategy revision. Messages and tool responses are condensed, not verbatim; identifiers are shortened. Each version generates 50 seeds. Negative feedback compares the addition with the preceding 2,050-seed pool; revision feedback compares equal-size 2,100-seed pools against one frozen baseline. Intermediate trials are indicated.}
  \label{tab:strategy-case-study}
  \small
  \setlength{\tabcolsep}{6pt}
  \renewcommand{\arraystretch}{1.12}
  \definecolor{caseAgent}{RGB}{34,63,98}
  \definecolor{caseTool}{RGB}{39,104,145}
  \definecolor{caseHarness}{RGB}{136,96,24}
  \definecolor{caseNegative}{RGB}{166,51,65}
  \definecolor{casePositive}{RGB}{24,119,100}
  \begin{tabularx}{\textwidth}{@{}L{0.105\textwidth}Y Y@{}}
    \toprule
    \textbf{Stage} & \textbf{Agent Decision and Tool Call} & \textbf{Tool Response and Harness Feedback} \\
    \midrule
    \textbf{1. Create} &
    \textcolor{caseAgent}{\textbf{Agent.}} Existing methods cover lot tracking, wayfinding, and direct settlement. Test \emph{multi-hop allocation} with intermediate-node conservation and shared bottlenecks.\par
    \textcolor{caseTool}{\textbf{Calls.}} \texttt{submit\_strategy(operator="create")} registers \emph{Conserved Network Allocation}; \texttt{generate\_seeds} requests its batch.
    &
    \textcolor{caseTool}{\textbf{Return.}} Strategy \texttt{4b71e76d}; generation queued.\par
    \textcolor{caseHarness}{\textbf{Harness.}} 50 seeds completed, including cache-import allocation and exhibit-box routing; the pool grows from 2,050 to 2,100 seeds.\par
    Scenario Vendi: $64.9761\!\rightarrow\!64.9427$;\par
    workflow Vendi: $44.2204\!\rightarrow\!44.1408$.\par
    \textcolor{caseNegative}{\textbf{$\Delta D_c=-0.0334$, $\Delta D_w=-0.0796$.}}\par
    MCP-combination coverage nevertheless rises by 8.
    \\
    \midrule
    \textbf{2. Diagnose} &
    \textcolor{caseAgent}{\textbf{Agent.}} More tool-combination coverage did not improve semantic diversity. Inspect actual seeds and their nearest existing scenarios before changing the strategy.\par
    \textcolor{caseTool}{\textbf{Call.}} \texttt{list\_strategy\_seed\_neighbors}\par
    \texttt{(field="scenario", offset=1)}.\par
    The agent also inspects workflow neighbors and seed pages.
    &
    \textcolor{caseTool}{\textbf{Return.}} New seed: \emph{an exhibit handler routes prop boxes through a staging bay}. Nearest existing seeds include:\par
    (a) \emph{a curator maintains a coherent offline exhibit trail}, cosine $0.6091$;\par
    (b) \emph{a hobby exhibit needs successor demonstration routes}, cosine $0.6080$.\par
    \textcolor{caseAgent}{\textbf{Agent finding.}} Repeated kit, desk, and classroom settings also share an allocation-and-balance skeleton. Similarity prompts inspection; it does not prove duplication.
    \\
    \midrule
    \textbf{3. Redesign} &
    \textcolor{caseAgent}{\textbf{Agent.}} Replace the repeated allocation skeleton with distinct coordinate, topology, raster, and geometry dependencies. Retain source authority, usable outputs, and independent checks.\par
    \textcolor{caseTool}{\textbf{Calls.}} \texttt{modify\_strategy} tests a spatial atlas, then two source-handoff repairs; the original remains active during evaluation.
    &
    \textcolor{caseHarness}{\textbf{Intermediate feedback.}} The first two spatial trials improve scenario/workflow Vendi but lose 4 units of MCP coverage. Neither is selected.\par
    \textcolor{caseAgent}{\textbf{Focused revision.}} Return to the stronger atlas, change four source-placement briefs, and preserve the other 46.\par
    \textcolor{caseTool}{\textbf{Return.}} Candidate \texttt{6a8871f1}; 50-seed generation queued with 15 frozen resources.
    \\
    \midrule
    \textbf{4. Reassess} &
    \textcolor{caseAgent}{\textbf{Agent.}} Inspect realized spatial seeds and neighbor pairs. A coordinate-registration case now tests a mapping with held-out controls; a boundary-membership case checks persisted results against original geometry. These change operative dependencies, not only domain labels.
    &
    \textcolor{caseHarness}{\textbf{Harness.}} Equal-quota projected replacement; the pool remains at 2,100 seeds:\par
    scenario Vendi: $64.9427\!\rightarrow\!65.7242$;\par
    workflow Vendi: $44.1408\!\rightarrow\!44.4232$.\par
    \textcolor{casePositive}{\textbf{$\Delta D_c=+0.7815$, $\Delta D_w=+0.2823$.}}\par
    MCP coverage: $1,018\!\rightarrow\!1,018$; one size-four combination replaces another. This is net preservation, not coverage expansion.
    \\
    \midrule
    \textbf{5. Select} &
    \textcolor{caseAgent}{\textbf{Agent.}} Select the retained source-placement revision: it improves complementary spatial construction without net MCP-coverage loss. The earlier trials' larger semantic gains do not justify their coverage regression.\par
    \textcolor{caseTool}{\textbf{Call.}} \texttt{select\_strategy\_modification}\par
    \texttt{root\_strategy\_id="4b71e76d"};\par
    \texttt{strategy\_id="6a8871f1"}. &
    \textcolor{caseTool}{\textbf{Return.}} Selected version recorded for this root; alternatives remain archived.\par
    \textcolor{caseHarness}{\textbf{Continuation.}} The selected revision enters the accepted pool used for subsequent strategy exploration. Later creation resumes from the finalized population, not from an unaccepted projection.
    \\
    \bottomrule
  \end{tabularx}
\end{table*}

\section{Case Study: Multi-Strategy Evolution}
\label{sec:strategy-case-study}
\label{sec:strategy-case-trace}
Table~\ref{tab:strategy-case-study} illustrates how the evolution agent uses diversity feedback to diagnose limited novelty, revise a generation strategy, and decide which candidate to retain. The trace follows one strategy through creation, diagnosis, redesign, reassessment, and selection.

\noindentparagraph{\textbf{\upshape From feedback to diagnosis.}}
The agent initially identifies multi-hop allocation as a promising direction and creates a conserved-network allocation strategy. Its 50 generated seeds increase MCP-combination coverage but reduce both scenario and workflow Vendi in the accumulated pool. This discrepancy shows why a plausible strategy description or a new tool combination alone is insufficient to establish semantic novelty. The agent inspects generated seeds and their nearest neighbors, identifying recurring settings and an allocation-and-balance workflow shared with existing tasks. Neighbor retrieval supplies concrete examples for diagnosis rather than treating the aggregate diversity decrease as a self-explanatory signal.

\noindentparagraph{\textbf{\upshape From diagnosis to revision.}}
The agent responds by changing the generation logic: spatial-computation variants introduce coordinate, topology, raster, and geometry dependencies in place of the repeated allocation structure. It then inspects the resulting seeds to check whether these changes produce substantively different operations, such as validating coordinate mappings with held-out controls and checking boundary membership against source geometry. Revision thus targets the structure of the generated work, not just scenario wording or domain labels.

\noindentparagraph{\textbf{\upshape Feedback-guided selection.}}
The agent does not simply select the candidate with the largest semantic-diversity gain. Earlier spatial trials improve scenario and workflow diversity but reduce MCP-combination coverage. The retained revision instead improves both semantic measures while preserving coverage under an equal-size replacement of the original batch. Once selected, this revision becomes part of the accepted pool against which subsequent exploration is evaluated. The case illustrates a concrete decision loop: aggregate feedback identifies a problem, seed-level comparisons inform a structural revision, and reassessment determines whether that revision should guide further synthesis.

This trace documents an extended-budget continuation with intermediate trials. Addition feedback compares the new pool with its preceding state, whereas revision feedback compares equal-size pools against a frozen baseline. It illustrates the agent's decision process rather than a controlled downstream-performance comparison; full messages, identifiers, and comparison baselines are preserved in the evidence archive.

\section{Artifact-Contrastive Rubric Construction and Verification}
\label{sec:rubric-implementation}
\label{sec:verification-protocol}
The prompts below are reproduced verbatim, with placeholders for task-specific inputs.

\noindentparagraph{\textbf{\upshape Rubric-author prompt and skill.}}
The invocation directs the author to the skill containing the rubric-construction instructions.

\begin{rubricprompt}{Rubric-Author Invocation}
Read and follow the rubric-author skill first:
{skill_path}

Task package:
{task_dir}

Fixed Teacher rollout directories:
{rollout_lines}

Write the task-specific rubrics to exactly one JSON file:
{output_path}

The host supplies fixed evaluator metadata. Do not create other artifacts.
\end{rubricprompt}

\begin{rubricprompt}{Rubric-Author Skill}
---
name: rollout-derived-rubric-author
description: Author task-grounded scoring rubrics from public initial state and fixed Teacher rollouts.
---

# Rollout-Derived Rubric Author

Write one `rubrics.json` that evaluates the task's final observable result. The task request and public initial state define what is required. Teacher rollouts are fallible evidence about how the task was attempted, not an answer key.

## Inspect And Decide

1. Read `task.md`, `initial_workspace/`, `manifest.json` when present, and any public task-provided service fixtures needed to understand the requested outcome.
2. Compare the final workspaces and other observable final states across the fixed Teacher rollouts. Read `conversation_history/messages.json`, `trajectory_events.jsonl`, and run logs when they help explain a disagreement or apparent omission.
3. For each substantive divergence, decide from the task and initial state which outcomes are correct, which are wrong, and which are acceptable alternatives. Different tool calls that produce equivalent final results do not need separate criteria.
4. Identify task requirements that none of the rollouts satisfied. Include those requirements in the rubrics as well. Agreement among the rollouts does not establish correctness.

Do this analysis while inspecting the evidence; do not write intermediate reports. The only required output is `rubrics.json`.
## Write Scorable Criteria

Describe required final states, not a preferred sequence of tool calls or a particular Teacher's behavior. A target may be a workspace file, repository state, database record, or other state the Judge can observe through the task's read-only tools and MCPs. Ground every criterion in the task request and public initial state; use rollout evidence to discover disagreements and omissions, not as a scoring dependency.

Use as many `rubric_items` as the distinct task outcomes require. Give each item a stable `id`, `type` (`exact` or `semantic`), positive `points`, concrete `target`, nonempty public `evidence` list, and a `check` stating what must hold. An item is binary: the Judge awards all its points only when the whole criterion is met. Split independently satisfiable outcomes into separate items. Weight the core requested result more heavily than superficial formatting or prose.

Use `exact` for concrete values, identities, schemas, calculations, and state transitions; use `semantic` where legitimate wording or judgment can vary. Both types are evaluated against observable state by the Agent-as-Judge, so an `exact` item must still say precisely what evidence establishes it. Accept alternatives allowed by the task, and never require a particular Teacher rollout or tool sequence.

`artifacts` and `protected_paths` are optional. Use an `artifact` only for an unambiguous required workspace file or structural condition, and a `protected_path` only when the task requires an initial file to remain byte-identical. These are global deterministic gates: a failure skips Judge scoring and fails the entire task. Do not use them for debatable business logic, acceptable alternatives, or external state. Put those requirements in `rubric_items` instead.

## Output

Write exactly one JSON object to the requested `rubrics.json` path. The task-specific authoring interface is:

```json
{
  "rubric_items": [
    {
      "id": "stable-id",
      "type": "exact",
      "points": 10,
      "target": "workspace:relative/output/path",
      "evidence": ["task.md", "initial_workspace/relative/source"],
      "check": "The required final state and any acceptable alternatives."
    }
  ]
}
```

Add `artifacts` or `protected_paths` only when an optional deterministic gate is justified. The host supplies the schema version, rollout-derived mode, pass threshold, and Judge configuration. Do not put rollout paths, messages, trajectories, model identities, or prior scores in `target`, `evidence`, or `check`.

Read `rubrics.json` back before finishing. Fix invalid JSON, unsupported fields, duplicate ids, vague targets, or criteria that cannot be judged from the final observable result.
\end{rubricprompt}

\noindentparagraph{\textbf{\upshape Final-state judge prompt.}}
The judge receives a task-specific evaluation packet and returns rubric verdicts, which the host converts into scores. Only valid trajectories receiving full rubric scores and passing all applicable hard checks are retained for training.

\begin{rubricprompt}{Final-State Judge Prompt}
You are a strict benchmark evaluator running inside a read-only Codex harness.
Judge only the final observable task state. Do not inspect rollout messages, conversation history, traj_log files, eval_res files, or model-specific traces. Do not modify the actual workspace, reference workspace, ground-truth workspace, or task files.

Use the task-specific read-only MCP tools and local read-only tools to inspect the actual workspace and compare it with task.md, the rubrics, the initial workspace, and any available oracle/reference materials. Some rubric-derived evaluators intentionally do not have a groundtruth_workspace; in that case, judge from the task semantics, public initial state, and observable final state. Inspect the underlying state directly: for repositories this includes read-only commands such as git status, git diff, git log, and git rev-parse; for other tasks use the available readers or query tools. When running a Python-based verification command, use the absolute python_executable from the evaluation packet rather than assuming `python` is on PATH. A report's claim is not evidence that the claimed state exists. If task.md contains {fixture_port}, use the port in fixture_base_url to inspect the task's local site.

Decide every rubric item independently with exactly one verdict: met, not_met, or unverifiable. Do not assign a numeric score and do not award partial credit. The evaluator owns the point values and computes the score deterministically. Use met only when direct observable evidence establishes the whole criterion. Use not_met when any required part is absent, wrong, contradictory, or incomplete. Use unverifiable only when the allowed state and tools cannot establish either met or not_met; do not use it as a substitute for inspection. Items marked exact describe deterministic facts, values, identities, schemas, calculations, or state transitions; judge them by the observable workspace state, not by brittle phrase matching. Items marked semantic describe narrative quality, acceptable paraphrases, and justifications. Do not reward unsupported claims or behavior that only appears in a trajectory.

Write exactly one JSON object to the judge_result_path. The file must have this schema:
{"rubric_results":[{"id":"...","verdict":"met|not_met|unverifiable","rationale":"...","evidence_used":["relative/or/absolute/path: short note"]}],"summary":"..."}

Evaluation packet:
{evaluation_packet}
\end{rubricprompt}

\section{Benchmark Details and Evaluation Protocols}
\label{sec:evaluation-score-details}

Scores are reported on a 0--100 scale; success rates are percentages.

\noindentparagraph{Toolathlon-Verified.}
Toolathlon-Verified contains 108 tasks requiring long-horizon tool use across applications, with specified initial states and task-specific outcome evaluators \cite{li2025toolathlon,toolathlon2026leaderboard}. Tasks involve information retrieval, dependent operations, and application-state updates. Pass@1 measures average single-run success; Pass@3 and Pass$^3$ measure success in at least one and all three runs, respectively.

\noindentparagraph{AutomationBench.}
AutomationBench contains 600 business-workflow tasks, with 100 each in sales, marketing, operations, customer support, finance, and human resources \cite{shepard2026automationbench}. Tasks require policy-compliant actions through REST APIs across simulated SaaS applications. We report average task score for graded performance and success rate for complete-task success.

\noindentparagraph{Job-Bench.}
Job-Bench contains 130 tasks across 35 occupations, targeting workflows prioritized by domain experts for delegation \cite{li2026jobbench}. Each task pairs heterogeneous reference files with a professional deliverable request. We evaluate the 65-task main split using OpenCode and report the aggregate score from task-specific, fact-anchored rubrics with binary criteria.

\noindentparagraph{WorkBuddy Bench (Office).}
The Office subset contains 50 tasks pairing user requests with sandboxed workspaces of spreadsheets, CSV files, PDFs, and Word documents \cite{tencent2026workbuddy}. Agents process these materials to produce office artifacts. We report the aggregate Office score, combining task-specific rule checks and semantic rubrics, rather than an all-checks pass rate.

\end{document}